\documentclass[journal]{IEEEtran}

\ifCLASSINFOpdf
\else
   \usepackage[dvips]{graphicx}
\fi
\usepackage{url}

\usepackage{algorithm}
\usepackage{algorithmic}
\usepackage{graphicx, cite}
\usepackage{multirow, booktabs, makecell}
\usepackage[table]{xcolor}
\definecolor{gray}{rgb}{0.85,0.85,0.85}
\usepackage{footnote}
\makesavenoteenv{tabular}
\makesavenoteenv{table}

\begin{document}

\title{Fidelity Estimation Improves Noisy-Image Classification with Pretrained Networks}

\author{Xiaoyu Lin, Deblina Bhattacharjee, Majed El Helou, \IEEEmembership{Members, IEEE}, and Sabine Süsstrunk, \IEEEmembership{Fellow, IEEE}. 
\thanks{Submitted for review on June 1, 2021. \\
This work was supported in part by the Swiss National Science Foundation via the Sinergia grant number CRSII5$-$180359.
}
\thanks{School of Computer and Communication Sciences, EPFL, Switzerland. \{xiaoyu.lin, deblina.bhattacharjee, majed.elhelou, sabine.susstrunk\} @epfl.ch}
}

\markboth{Published in IEEE Signal Processing Letters}
{}
\maketitle
\begin{abstract}
Image classification has significantly improved using deep learning. This is mainly due to convolutional neural networks (CNNs) that are capable of learning rich feature extractors from large datasets. However, most deep learning classification methods are trained on clean images and are not robust when handling noisy ones, even if a restoration preprocessing step is applied. 
While novel methods address this problem, they rely on modified feature extractors and thus necessitate retraining. 
We instead propose a method that can be applied on a \textit{pretrained} classifier. Our method exploits a fidelity map estimate that is fused into the internal representations of the feature extractor, thereby guiding the attention of the network and making it more robust to noisy data. 
We improve the noisy-image classification (NIC) results by significantly large margins, especially at high noise levels, and come close to the fully retrained approaches. Furthermore, as proof of concept, we show that when using our oracle fidelity map we even outperform the fully retrained methods, whether trained on noisy or restored images. 
\end{abstract}


\IEEEpeerreviewmaketitle

\section{Introduction} \label{sec:intro}
Traditionally, image classification methods have relied on a limited set of hand-crafted features~\cite{boiman2008defense, yang2009linear, perronnin2010improving, kobayashi2013bfo, lowe2004distinctive, dalal2005histograms}. Deep learning approaches, especially supervised methods using convolutional neural networks (CNNs)~\cite{griffin2007caltech,krizhevsky2012imagenet} trained on large annotated datasets~\cite{deng2009imagenet}, are capable of extracting rich feature representations~\cite{deng2009imagenet,schlegl2014unsupervised} and thus improve performance. However, obtaining reliable and generalizable feature extractors remains a challenge~\cite{elhelou2020al2,li2019understanding,blot2018shade,morcos2018importance}. It is thus desirable to exploit 
existing, pretrained feature extractors in a modular way across datasets or applications without the need to retrain them. Here, we exploit the feature extractor pretrained on \textit{clean} images to work with \textit{noisy} input.

{\let\thefootnote\relax\footnotetext{All code and supplementary material at \url{https://github.com/IVRL/FG-NIC}}}

Classification methods trained on datasets with clean images show a significant performance drop on noisy or restored images~\cite{tian2020deep}. Yet, in many applications, the captured images are inherently noisy, for instance, in low-light conditions, short exposure capture, in multi-spectral acquisition, and when employing inexpensive surveillance cameras, etc. A recent study~\cite{pei2019effects} confirmed that image classification accuracy drops significantly when the input images are noisy. Furthermore, even a straight-forward restoration preprocessing step failed to improve classification.  

Recently, two studies specifically addressed Noisy-Image Classification (NIC). However, they either require costly retraining~\cite{borkar2019deepcorrect} or even necessitate an entire redesign of the CNN-based classifier~\cite{Li_2020_CVPR}. 

\begin{figure}
    \centering
    \includegraphics[width=.95\linewidth]{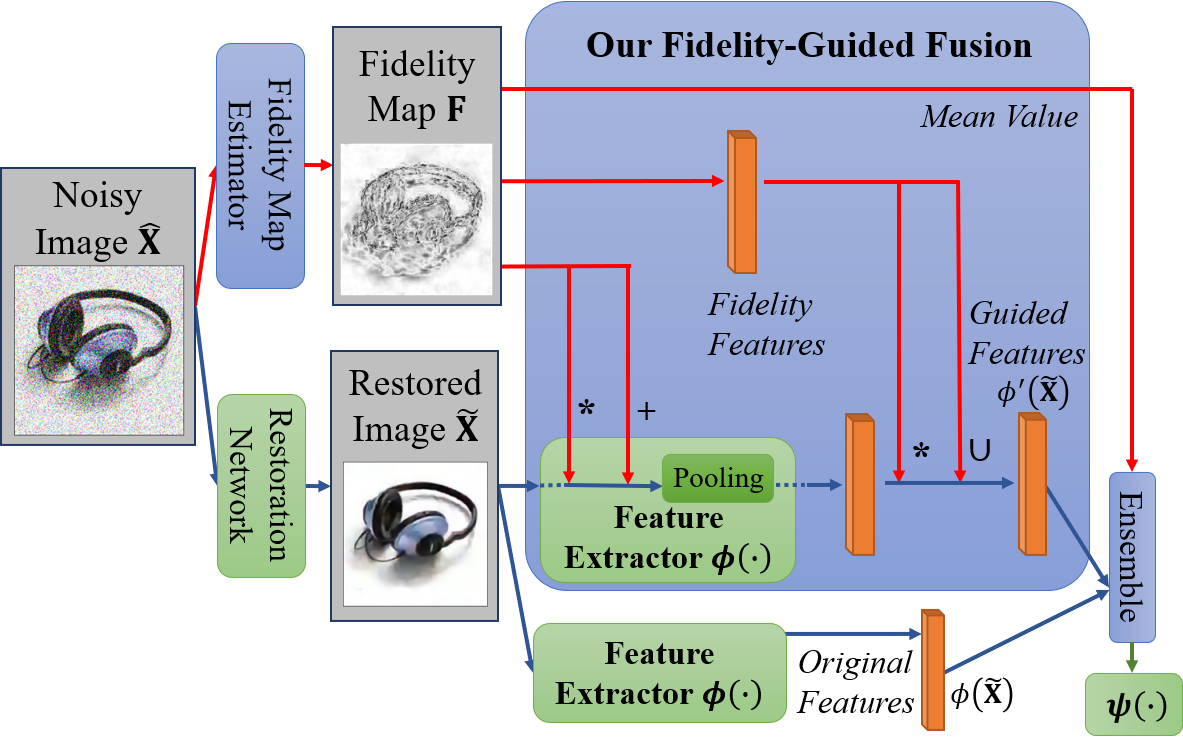}
    \caption{High-level overview of our FG-NIC method. 
    During training, we freeze the pretrained classification and restoration networks, highlighted with green blocks. The trainable modules are highlighted with blue blocks. (*), (+), ($\cup$), and $\psi(\cdot)$ denote multiplication, addition, concatenation and the fully connected part of a classification network, respectively.}
    \label{fig:pipeline}
\vspace{-10pt}
\end{figure}

In contrast, our proposed approach can be extended to any feature extractor \textit{without} retraining, enabling existing networks to also classify noisy images. Our Fidelity-Guided Noisy-Image Classification (FG-NIC) method exploits an internal integration of a data fidelity estimator within the feature extractor to adjust the features at test time. We estimate the fidelity map with a deep network. We then exploit the fidelity map to guide the attention of the pretrained feature extractor by fusing its weights into the different network layers (see Fig.~\ref{fig:pipeline}). Experimental evaluations show that our method improves the baseline by large margins and even rivals fully retrained classifiers. Furthermore, our proof-of-concept method with an oracle fidelity map outperforms even all the fully retrained classification networks.

\section{Related work}\label{sec:related}
Deep networks outperform human classification on clean images~\cite{tadros2019assessing}. However, their performance deteriorates on noisy images on which humans do better~\cite{tadros2019assessing}. A recent study~\cite{pei2019effects} demonstrates the negative effect of degraded images, including hazy images, underwater images, and motion-blurred images, on classification accuracy. Their results parallel those of an earlier work~\cite{roy2019effects}. Both studies emphasize the need to develop robust NIC methods. 
Validating these observations, two recent approaches, DeepCorrect~\cite{borkar2019deepcorrect} and WaveCNets~\cite{Li_2020_CVPR}, address the problem of NIC under additive white Gaussian noise (AWGN). The authors of DeepCorrect~\cite{borkar2019deepcorrect} analyze how feature extractor activations are affected by noise. 
To do so, DeepCorrect estimates how susceptible each filter is to the AWGN degradation and learns to replace these filters by retraining them. WaveCNets proposes a wavelet-decomposition network architecture~\cite{Li_2020_CVPR}. As noise mostly affects the high-frequency components of an image, WaveCNets trains a more noise-robust network that focuses on low frequencies. Both these methods need retraining and thus do not benefit from pretrained classifiers and feature extractors. 

The most recent NIC method proposes an ensemble network consisting of a restoration network followed by two classification ones for clean and for degraded images~\cite{endo2020classifying}. Though requiring retraining, this method does exploit one pretrained network for some \textit{clean} images. To be modular and be able to deploy novel pretrained classifiers despite noise, our solution exploits \textit{only pretrained} classifiers. This way, we avoid the need for computationally expensive retraining, and can directly make use of future, potentially improved, classifiers.

Another way to approach NIC is to apply a restoration preprocessing step. However, when preprocessing with a restoration network, for instance the DnCNN~\cite{zhang2017beyond} denoiser, the performance of the classification network is still severely affected, which we show in our experimental section. There is an observable trade-off between the perceptual quality of the restoration output and the classification performance that was recently highlighted in~\cite{NEURIPS2019_6c29793a}. For that reason, we rely on a fidelity map that we can either learn explicitly or, as we show in our supplementary material, learn internally in an end-to-end approach~\cite{elhelou2020bigprior}. We use this data fidelity to guide the attention of the network~\cite{fu2020scene,ma2021deep} and account for potential distortions coming from the degradation or from the restoration steps. Learning an internal indicator for various degradation levels was recently shown to improve the performance and robustness of deep denoisers~\cite{elhelou2020blind}. In our case, we use fidelity as an attention guide for adjusting the activations of the feature extractor in a classifier.



\section{Method}
\label{sec:method}
We present our proposed FG-NIC method by first introducing our fidelity map that we fuse into the pretrained classifier. We discuss the high-level manipulation of the pretrained classifier 
(with fixed weights) 
and present the details of our fidelity fusion.
The overall pipeline is shown in Fig.~\ref{fig:pipeline}. An extended figure with the specific fidelity fusion details is in the supplementary material. 

\subsection{Fidelity map}\label{subsec:fidelity}
A fundamental element of our approach is the fidelity map that we use to adjust the feature extractor's output and to guide the attention of the network. Given an image restored by a pretrained network, some image components are more faithful to the original clean image than others. This can depend on the noise (possibly spatially varying), the frequency components or the texture across the image, and on the pretrained restoration network itself. We define the fidelity map in terms of the element-wise $\ell_1$ distance between the original clean image $\mathbf{X}$ and the restored image $\widetilde{\mathbf{X}}=R(\hat{\mathbf{X}})$, where $R(\cdot)$ is the restoration network and $\hat{\mathbf{X}}$ is the noisy image corresponding to $\mathbf{X}$.
In the supplementary material, we further test fidelity maps based on $\ell_2$ and cosine distances. While the $\ell_2$ distance performs comparably to the $\ell_1$ distance for lower noise levels, the $\ell_1$ distance outperforms it at higher noise levels, due to its robustness to outliers. We hence rely on a fidelity map based on the $\ell_1$ distance. 
This fidelity map at pixel $(i,j)$ is
\begin{equation}
    \mathbf{F}_{(i,j)}=1-\frac{1}{C}\sum^C_{c=1}|\widetilde{\mathbf{X}}_{(i,j,c)}-\mathbf{X}_{(i,j,c)}|
    \label{eq:fidelity}
\end{equation}
where $\widetilde{\mathbf{X}}_{(i,j,c)}$ is the pixel intensity of the restored image at pixel $(i,j)$ and channel $c$, and similarly $\mathbf{X}_{(i,j,c)}$ is the pixel intensity of the clean image $\mathbf{X}$, and $C$ is the number of channels (3 for RGB).
All images and the fidelity map are defined over $[0, 1]$. From our definition for the fidelity map, it indicates our confidence level in each restored pixel. 
The ground-truth oracle fidelity map is naturally not available at test time. Therefore, we train a deep network to act as the fidelity map estimator for the deep restoration network that we use. In our experiments, we implement the fidelity estimator using a state-of-the-art residual architecture~\cite{zhang2017beyond}. 

\subsection{Manipulated classifier}
For a given CNN-based image classification network $f(\cdot)$, we refer to its {feature extractor} part that outputs the high-level features for an image as $\phi(\cdot)$. And we refer to its {feature classifier}, which mainly consists of fully connected layers for predicting the class likelihoods, as $\psi(\cdot)$. 
We formulate our method as a manipulation over the \textit{pretrained} feature extractor $\phi(\cdot)$. As we discuss in detail in the following section, we exploit our fidelity map to guide the attention of the pretrained feature extractor by fusing its information into the different network layers. This enables us to adapt the importance that the network assigns to different features based on the varying quality of the degraded-then-restored image. The manipulated classification network can then be formulated generally as $\bf{y} = \psi(\phi'(\widetilde{\bf{X}}))$ 
where $\phi'(\widetilde{\bf X})$ is the output of the manipulated feature extractor, relative to the original extractor $\phi(\widetilde{\bf X})$, and $\psi(\cdot)$ are the fully connected layers that output the class likelihoods $\bf{y}$. Our approach enables us to rely on $\phi(\cdot)$ and $\psi(\cdot)$ taken directly from a classification network that is pretrained on clean images. Hence, 
without retraining the pretrained classifier
, we can integrate a novel classification network in a straight-forward manner by obtaining $\phi'(\cdot)$ as described in the following section.

\subsection{Fidelity fusion}
To obtain the manipulated feature extractor, $\phi'(\cdot)$, relative to the original pretrained extractor $\phi(\cdot)$, we build our fusion over $\phi(\cdot)$ and describe our Algorithm~\ref{alg:fidelity_fusion} in what follows.  

\textbf{Spatial linear fusion.} 
The input restored image $\widetilde{\mathbf{X}}$ is affected by the degradation and restoration process. Even when noise is spatially uniform, different spatial regions in an image are affected to varying degrees based on their frequency content~\cite{el2020stochastic}. The objective of our fidelity map is thus to guide the attention of the feature extractor accordingly. To that end, we apply an element-wise spatial linear fusion on feature layers using our fidelity map weight. 
This translates into an element-wise multiplication and addition of the features with the fidelity map. This mapping is extended across all the feature channels of the pretrained network.
We add a trainable block 
(whose weights are learnt) 
between two sequential spatial fusion operations to increase the flexibility of our fusion. 

\begin{algorithm}[t]
\small
\caption[FG-NIC: Fidelity fusion algorithm.]{FG-NIC: Fidelity fusion.\footnotemark}
\label{alg:fidelity_fusion}
\begin{algorithmic}[1]
    \REQUIRE{{$\widetilde{\mathbf{X}}$ \# Input restored image}}
    \REQUIRE{$\mathbf{F}$ \# Our fidelity map}
    \REQUIRE{${\phi(\cdot) = Layer^N(...(Layer^1(\cdot)))}$ \# Pretrained extractor}
    \REQUIRE{$[Block^0_{m}(\cdot),...,Block^T_{m}(\cdot)]$ \# CNN for multiplication}
    \REQUIRE{$[Block^0_{a}(\cdot),...,Block^T_{a}(\cdot)]$ \# CNN for addition}
    \REQUIRE{$Block_{f}(\cdot)$ \# FC block for channel fusion}
    \REQUIRE{$Block_{c}(\cdot)$ \# FC block for channel concatenation}
        \item[] \# Spatial linear fusion
        \STATE $\mathbf{f}_m \gets Block^0_{m}(\mathbf{F})$;
                $\mathbf{f}_a \gets Block^0_{a}(\mathbf{F})$;
                $t \gets 1$
        \STATE $\mathbf{x} \gets (2\times Sigmoid(\mathbf{f}_m) \otimes\widetilde{\mathbf{X}}) \oplus \mathbf{f}_a$
    \FOR{$n = 1,...,N$}
        \IF{$Layer^n(\cdot)\in$ PoolingLayers}
            \STATE $\mathbf{f}_m\gets Down Sampling(Block^t_{m}(\mathbf{f}_m) + \mathbf{f}_m)$
            \STATE $\mathbf{f}_a\gets Down Sampling(Block^t_{a}(\mathbf{f}_a) + \mathbf{f}_a)$
            \STATE $\mathbf{x} \gets (2\times Sigmoid(\mathbf{f}_m) \otimes\mathbf{x}) \oplus \mathbf{f}_a$
            \STATE $t\gets t+1$
        \ENDIF
            \STATE $\mathbf{x} \gets Layer^n(\mathbf{x}) $
    \ENDFOR
    
    \item[] \# Channel  fusion
    \STATE $\mathbf{f}_c \gets Down Sampling (Flatten(\mathbf{F}))$
    \STATE $\mathbf{x} \gets 2\times Sigmoid(Block_{f}(\mathbf{f}_c)) \circ\mathbf{x}$
    \item[] \# Channel  concatenation
    \STATE $\mathbf{x} \gets Block_c(Concatenate(\mathbf{f}_c, \mathbf{x}))$
    \ENSURE Modified feature $\mathbf{x}=\phi'(\widetilde{\mathbf{X}})$ 
\end{algorithmic}
\end{algorithm}
\footnotetext{$\circ$ is the Hadamard product, $\otimes$ the Hadamard product spatially with broadcasting to all channels, and $\oplus$ the element-wise addition in spatial dimensions with channel broadcasting. FC refers to fully-connected layers.}

\begin{figure*}
    \centering
    \includegraphics[width=0.95\linewidth]{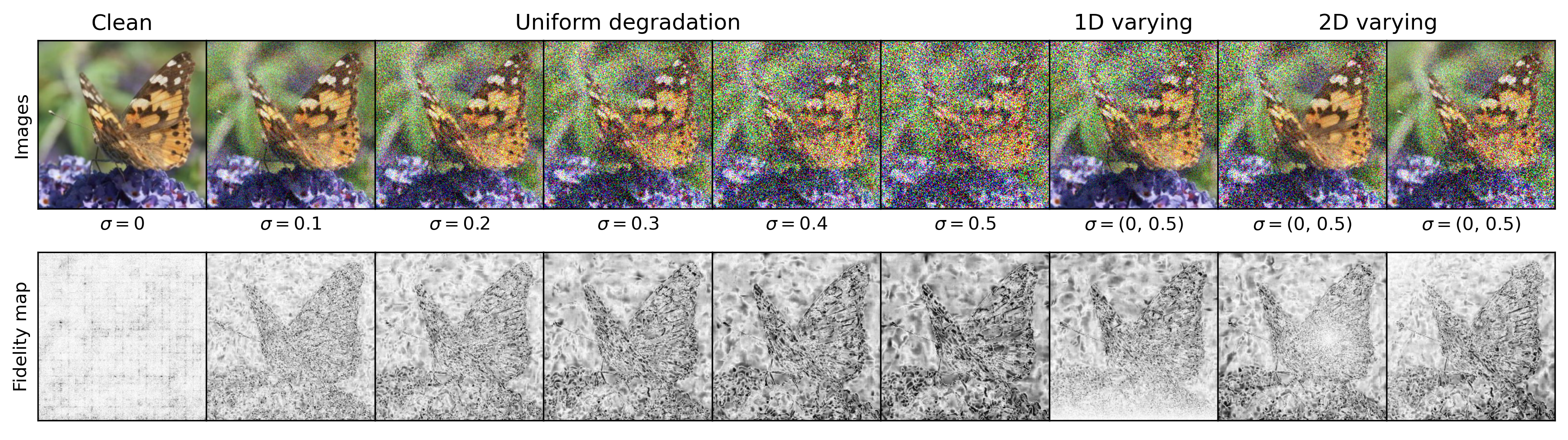}
    \vspace{-0.5cm}
    \caption{Illustrative example with images degraded by AWGN, from the \textit{'butterfly'} category in Caltech-256. For the spatially varying noise, $\sigma=(0, 0.5)$ indicates that the degradation level varies linearly from $0$ to $0.5$. The bottom row shows our fidelity maps averaged over the RGB channels, corresponding to each image in the top row. We apply a gamma correction to the fidelity maps for better visualization. }
    \label{fig:images}
\vspace{-10pt}
\end{figure*}

\textbf{Channel fusion.}
The aforementioned modules are all applied spatially. However, the feature extractor also stores information across channels. For example, the output of the ResNet-50~\cite{he2016deep} feature extractor has 2048 channels. For that reason, it is important to also guide the channel attention of the feature extractor.
We could, in theory, train a channel-based fidelity map with 2048 channels, however, the complexity and computational cost would significantly increase.
We thus flatten and downsample our fidelity map to obtain a new fidelity feature having the same length as the pretrained network's feature space.
As with the spatial fusion, the features we obtain by manipulating and reshaping our fidelity map are passed through a trainable block 
(whose weights are learnt), 
for more flexibility. We fuse the output of this block with the image features through channel-wise multiplication, to guide the attention of the feature extractor across channels.

\textbf{Channel concatenation.} We improve the flexibility of our fusion method by preserving the fidelity information and enabling a trainable block to perform further integration of both the fidelity map and the image features. We concatenate our fidelity map and the image features together and pass the concatenated features (of size 2$\times$2048 for the ResNet-50 architecture) into a trainable block. To enable the use of the pretrained network, the resulting feature set retains the same size as the original one.

\textbf{Ensembling.} We also investigate using an ensemble of {$\{\phi(\cdot)$, $\phi'(\cdot)\}$} rather than only our manipulated features. Our ensemble module assigns element-wise weights to the two feature layers to combine them. This approach enables our method to exploit the best of the two feature spaces, without having to modify the pretrained fully connected layers. As we show in our experimental results, preserving the original features can be beneficial at low degradation levels.

\begin{table}
  \setlength\tabcolsep{3.5pt}
  \centering
  \caption{Top-1 accuracy (\%) of state-of-the-art approaches and our FG-NIC for noisy images, averaged over all noise levels, on Caltech-256. The best and second best results \emph{aside from the oracle configuration} are in bold and underlined respectively.
  }
    \begin{tabular}{lccc}
    \toprule
    Methods    & AlexNet & ResNet-18 & ResNet-50 \\
    \midrule
    Pretrained (test on noisy) & 13.08 & 22.11 & 23.27\\
    
    Pretrained (test on restored~\cite{zhang2017beyond}) & 44.42 & 46.15 & 52.33 \\
    
    Retrain on noisy & \underline{52.87} & \bf{65.54} & \bf{72.95} \\
    
    WaveCNets~\cite{Li_2020_CVPR} (Haar)~\footnotemark & - & 22.94 & 24.68\\
    
    DeepCorrect~\cite{borkar2019deepcorrect} & 21.90 & 42.14 & 45.54\\
    
    FG-NIC (pretrained + single)  & 44.53 & \underline{57.99} & \underline{68.69} \\
    
    FG-NIC (pretrained + ensemble)  & \bf{54.61} & 56.99 & 67.04 \\
    
    
    
    \midrule
    FG-NIC (oracle + single)  & \textit{57.49} & \textit{68.22} & \textit{76.10}\\
    FG-NIC (oracle + ensemble)  & \textit{58.51} & \textit{68.59} & \textit{75.42}\\
    
    \bottomrule
    \end{tabular}%
  \label{tab:baseline}%
\vspace{-10pt}
\end{table}%
\footnotetext{Pretrained models: {https://github.com/LiQiufu/WaveCNet}. Note, we do not report the AlexNet perfomance on the WaveCNets model as the authors do not provide the pretrained model.}

    
    
    
    
    
    
    
    

\section{Experiments} \label{sec:experiments}

\subsection{Data processing}
We use all categories of the Caltech-256 dataset~\cite{griffin2007caltech}, and perform the standard processing and augmentation used in the state-of-the-art methods~\cite{pei2019effects,he2019bag}. 
For our AWGN degradation, the additive noise is independent across color channels. 
For different degradation levels, we change the standard deviation ($\sigma$) of the Gaussian noise. For further evaluation, we implement spatially varying degradation with 1D and 2D variations. The noise level linearly changes with the number of rows or columns for 1D varying noise, and for 2D varying noise, the noise level linearly changes with respect to the Euclidean distance from a random point on the image.
Fig.~\ref{fig:images} illustrates some examples with different AWGN degradations on an image along with their corresponding fidelity maps.

\vspace{-5pt}
\subsection{Baseline methods}
Aside from the pretrained baseline, we compare against multiple fully retrained networks. Each of our setups is then tested on images with and without restoration preprocessing. 
In the \textbf{Pretrained} baseline, the classification network is only trained on clean images, and we do not modify it. 
We also retrain the classification network on a mixture of noisy and clean images (\textbf{Retrain on noisy}), i.e., $\sigma\in\{0, 0.1, 0.2, 0.3, 0.4, 0.5\}$. For the \textbf{Retrain on restored} baseline, the training set consists of the outputs of the restoration network. The restored images are obtained by processing the same images as in the previous setup with the DnCNN denoiser~\cite{zhang2017beyond}.

Our proposed \textbf{FG-NIC} method, as presented in Sec.~\ref{sec:method}, exploits a fidelity map to manipulate the image features obtained by a pretrained classification network. The pretrained classifier is the same network used in the first \textbf{Pretrained} baseline, along with a pretrained preprocessing denoiser~\cite{zhang2017beyond}. We use a pretrained fidelity estimator, which we obtain with the same training setup as that of the denoiser~\cite{zhang2017beyond}. 
In our FG-NIC method we also design a proof-of-concept configuration that uses an \textbf{oracle} fidelity map. This is the ground-truth fidelity map, computed following Eq.~\ref{eq:fidelity}, that we use instead of an estimated fidelity map. This is to study the performance of our proposed approach independent of the quality of the fidelity map estimator. 
As an ablation study, we allow the network to fine-tune its own learned fidelity estimation with end-to-end training, and present these extended results in the supplementary material. We also provide in the the supplementary material, the number of operations and parameters used by the different approaches.
\vspace{-5pt}
\subsection{Experimental results}
\begin{table}
  \setlength\tabcolsep{2pt}
  \centering
  \caption{Classification accuracy (\%) of the baselines and our FG-NIC methods (based on ResNet-50) on Caltech-256. The best and second best results \emph{aside from the oracle configuration} are in bold and underlined respectively.}
    \begin{tabular}{lllccccccc}
    \toprule
    \multirow{2}{*}{Methods} & \multicolumn{2}{c}{\multirow{2}{*}{\makecell[c]{Experimental\\ setup}}} & \multicolumn{5}{c}{Uniform degradation ($\sigma$)} &  \multicolumn{2}{c}{Varying} \\
    
    \multicolumn{3}{c}{}  
    & 0.1   & 0.2   & 0.3   & 0.4   & 0.5   & 1D    & 2D \\
    \hline
    \noalign{\smallskip}
    
    \multirow{2}{*}{Pretrained} 
    & \multicolumn{2}{l}{\cellcolor{gray}Test on noisy} 
    & 67.60 & 32.68 & 11.25 &  3.50 &  1.35 & 30.25 & 27.15 \\
             
    & \multicolumn{2}{l}{\cellcolor{gray}Test on restored} 
    & 77.99 & 67.06 & 53.00 & 38.08 & 25.53 & 64.08 & 62.55 \\
    \noalign{\smallskip}
    
    \multirow{2}{*}{\makecell[l]{Retrain on\\noisy}} 
    & \multicolumn{2}{l}{\cellcolor{gray}Test on noisy} 
    & 79.17 & \underline{76.21} & \underline{73.08} & \underline{69.90} & \underline{66.39} & \underline{74.71} & \underline{74.71} \\
    
    & \multicolumn{2}{l}{\cellcolor{gray}Test on restored} 
    & 78.44 & 74.14 & 67.47 & 59.15 & 50.15 & 71.69 & 71.43 \\ 
    \noalign{\smallskip}
    
    \multirow{2}{*}{\makecell[l]{Retrain on\\restored}} 
    & \multicolumn{2}{l}{\cellcolor{gray}Test on noisy} 
    & 71.33 & 46.07 & 20.02 &  7.43 &  2.26 & 42.32 & 39.06 \\
                  
    & \multicolumn{2}{l}{\cellcolor{gray}Test on restored} 
    & \bf{80.32} & \bf{77.85} & \bf{75.04} & \bf{71.90} & \bf{68.37} & \bf{76.88} & \bf{76.70} \\
    \noalign{\smallskip}

    \multirow{2}{*}{\makecell[l]{FG-NIC\\(Pretrained)}} & \multicolumn{2}{l}{\cellcolor{gray}Single}
    & \underline{79.57} & 74.90 & 69.37 & 63.19 & 56.44 & 73.29 & 72.87 \\
    
    & \multicolumn{2}{l}{\cellcolor{gray}Ensemble}
    & \bf{80.32} & 74.84 & 68.18 & 60.31 & 51.56 & 73.34 & 72.68 \\
    
    \midrule

    \multirow{2}{*}{\makecell[l]{FG-NIC\\(Oracle)}} & \multicolumn{2}{l}{\cellcolor{gray}Single}
    & \textit{80.45} & \textit{78.14} & \textit{76.13}& \textit{74.06} & \textit{71.75} & \textit{77.64} & \textit{77.52} \\
    
    & \multicolumn{2}{l}{\cellcolor{gray}Ensemble}
    & \textit{81.14} & \textit{78.28} & \textit{75.52} & \textit{72.64} & \textit{69.51} & \textit{77.77} & \textit{77.41} \\

    \bottomrule
    \end{tabular}%
  \label{tab:experiment}%
  \vspace{-10pt}
\end{table}%

We compare the accuracy of the state-of-the-art methods DeepCorrect~\cite{borkar2019deepcorrect} and WaveCNets~\cite{Li_2020_CVPR} with our FG-NIC on noisy images. The results in Table~\ref{tab:baseline} are averaged over all uniform noise levels (0.1 to 0.5). We evaluate the models with AlexNet, ResNet-18, and Resnet-50. We report the performance of pretrained models, pretrained models with restoration preprocessing, retrained models, WaveCNets, and DeepCorrect. We observe that our FG-NIC (pretrained), outperforms all other models, and comes close to the one retrained on noisy data. Our FG-NIC (oracle), which is the proof-of-concept method, consistently reports the best performance.

In Table~\ref{tab:experiment}, we show classification accuracy on various levels of image noise, both uniform and spatially varying. Our methods significantly outperform the pretrained baseline, whether the latter is tested on noisy or restored images, consistently over all noise levels. Comparing with the fully retrained networks, we note that the accuracy of our method comes close, even though the gap increases at high noise levels. Again, our proof-of-concept method with the ground-truth oracle fidelity map outperforms even the fully retrained networks across all noise levels.
We also observe that although our ensemble module can slightly boost the classification performance at low noise levels, it quickly becomes detrimental to the results at higher levels. This indicates that our fidelity-based manipulated features $\phi'(\widetilde{\bf{X}})$ are better than the original ones when noise is present. Interestingly, the retrained baselines also tend to overfit their training distribution and are consequently not robust. For instance, the network retrained on noisy images performs significantly worse when given restored images, and the network retrained on restored images performs significantly worse when given noisy images. 

\section{Conclusion}\label{sec:ccl}
We propose the first method that manipulates the layers of a feature extractor to enable the classification of noisy images by a \textit{pretrained} classifier. By internally fusing data fidelity information, our FG-NIC model guides the attention of the classifier to improve its robustness to noise. We thus avoid retraining a 
pretrained classification network 
and still significantly improve over the baseline. With our oracle proof-of-concept approach we even consistently outperform, over all noise levels, the fully retrained classifier that is retrained specifically on the outputs of the restoration network. Our method can be readily extended to different degradation types in future work. It is also modular and can be used with future deep classifiers or restoration networks without the retraining overhead.\\


\bibliographystyle{IEEEtran}
\bibliography{references}

\end{document}


\title{Fidelity Estimation Improves Noisy-Image Classification with Pretrained Networks}

\author{Xiaoyu Lin, Deblina Bhattacharjee, Majed El Helou, \IEEEmembership{Members, IEEE}, and Sabine Süsstrunk, \IEEEmembership{Fellow, IEEE}.
}
\markboth{Supplementary Material}
{}
\maketitle

\IEEEpeerreviewmaketitle



\begin{figure}
\centering
    \includegraphics[width=\linewidth]{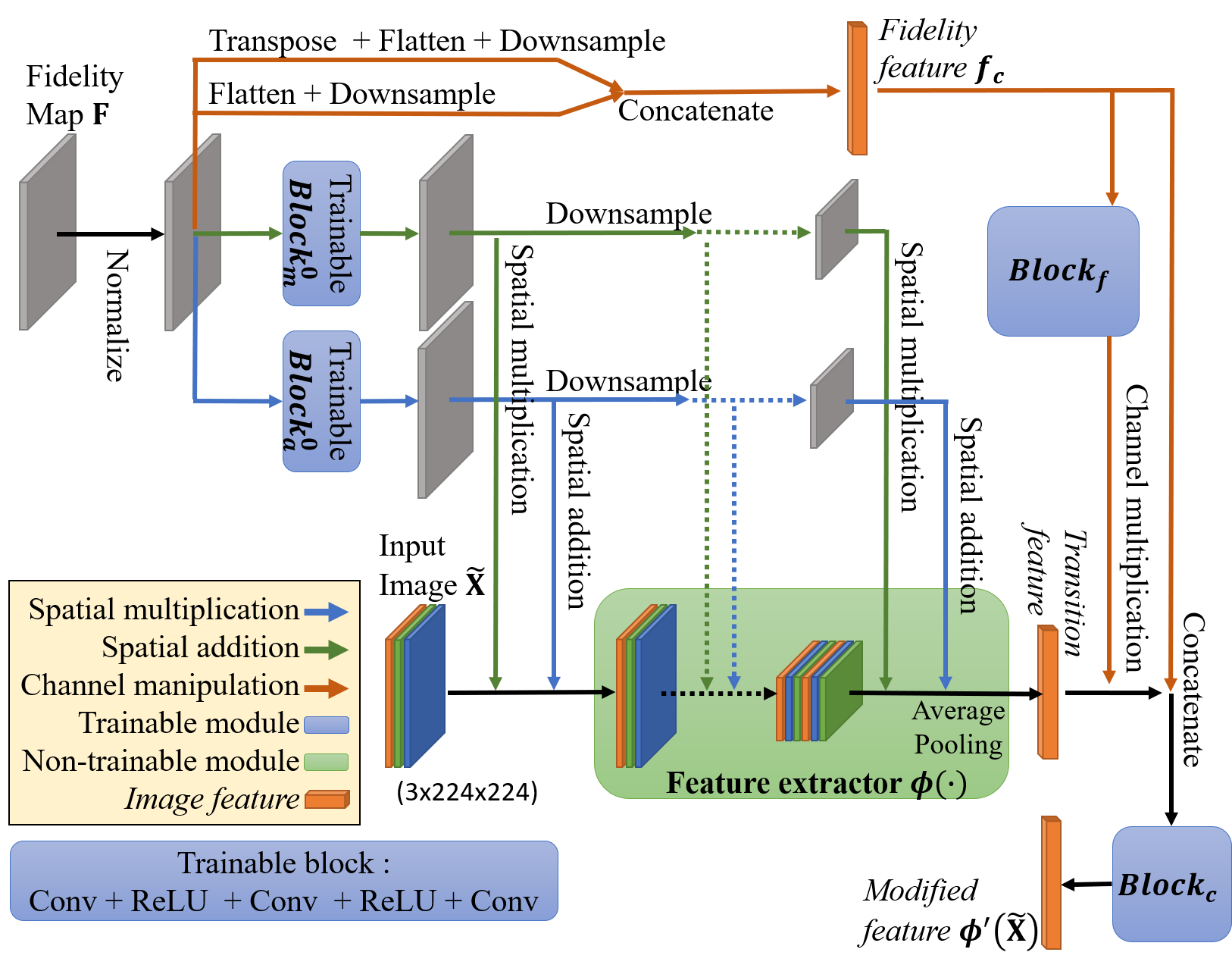}
    \caption{Detailed module architecture (without ensemble) of our proposed method based on ResNet-50. FC, Conv and ReLU indicate one fully connected layer, one convolutional layer and a rectified linear unit activation, respectively. The pretrained module (feature extractor) is shown in green, and the modules containing trainable parameters are shown in blue. 
    The pretrained module is the base classifier trained on clean images with fixed weights. The weights of the trainable blocks are learnt in our model.
    }
    \label{fig:details}
\end{figure}

\section{Extended experimental results}
The Caltech-101~\cite{fei2006one} dataset is similar to Caltech-256 with fewer images and categories. We select 30 images per class as the training set and keep the same procedure as for Caltech-256. The results are consistent and given in Table~\ref{tab:experiment}.
\begin{table}
  \setlength\tabcolsep{2pt}
  \centering
  \caption{Classification accuracy (\%) of the baseline and our FG-NIC methods (based on ResNet-50) on Caltech-101. The best and second best results \emph{aside from the oracle configuration} are in bold and underlined respectively.}
    \begin{tabular}{lllccccccc}
    \toprule
    \multirow{2}{*}{Methods} & \multicolumn{2}{c}{\multirow{2}{*}{\makecell[c]{Experimental\\ setup}}} & \multicolumn{5}{c}{Uniform degradation ($\sigma$)} &  \multicolumn{2}{c}{Varying} \\
    
    \multicolumn{3}{c}{}  
    & 0.1   & 0.2   & 0.3   & 0.4   & 0.5   & 1D    & 2D \\
    \hline
    \noalign{\smallskip}
    
    \multirow{2}{*}{Pretrained} 
    & \multicolumn{2}{l}{\cellcolor{gray}Test on noisy}
    & 83.87 & 56.95 & 27.13 & 10.95 &  5.50 & 54.23 & 50.18 \\
    
    & \multicolumn{2}{l}{\cellcolor{gray}Test on restored} 
    & \bf{92.03} & 84.83 & 70.16 & 50.17 & 31.18 & 81.38 & 80.17 \\
    \noalign{\smallskip}
    
    \multirow{2}{*}{\makecell[l]{Retrain on\\noisy}}
    & \multicolumn{2}{l}{\cellcolor{gray}Test on noisy} 
    & 89.62 & 87.92 & \underline{86.23} & \underline{84.50} & \underline{82.04} & 87.59 & 87.38 \\
    
    & \multicolumn{2}{l}{\cellcolor{gray}Test on restored} 
    & 89.85 & 87.90 & 82.16 & 73.98 & 62.19 & 85.86 & 85.68 \\ 
    \noalign{\smallskip}
    
    \multirow{2}{*}{\makecell[l]{Retrain on\\restored}} 
    & \multicolumn{2}{l}{\cellcolor{gray}Test on noisy} 
    & 84.18 & 64.52 & 36.82 & 15.53 &  5.85 & 59.62 & 56.74 \\
                  
    & \multicolumn{2}{l}{\cellcolor{gray}Test on restored} 
    & \underline{91.89} & \bf{89.86} & \bf{87.84} & \bf{85.39} & \bf{82.64} & \bf{89.49} & \bf{89.11} \\
    \noalign{\smallskip}
    
    \multirow{2}{*}{\makecell[l]{FG-NIC\\(Pretrained)}} 
    
    & \multicolumn{2}{l}{\cellcolor{gray}Single} 
    & 89.85 & 87.76 & 84.68 & 80.49 & 75.39 & 86.42 & 86.64 \\
    
    & \multicolumn{2}{l}{\cellcolor{gray}Ensemble} 
    & 91.79 & \underline{89.08} & 84.72 & 78.81 & 70.80 & \underline{87.71} & \underline{87.60} \\
    
    \midrule
    \multirow{2}{*}{\makecell[l]{FG-NIC\\(Oracle)}} 
    & \multicolumn{2}{l}{\cellcolor{gray}Single} 
    & \textit{90.91} & \textit{89.46} & \textit{87.75} & \textit{86.07} & \textit{83.83} & \textit{89.02} & \textit{88.89} \\
    
    & \multicolumn{2}{l}{\cellcolor{gray}Ensemble} 
    & \textit{91.97} & \textit{90.29} & \textit{88.02} & \textit{85.82} & \textit{82.87} & \textit{89.61} & \textit{89.53} \\
    
    
    
    \bottomrule
    \end{tabular}%
  \label{tab:experiment}%
\vspace{-10pt}
\end{table}%

For results on CUB-20-2011 dataset, please refer to (\url{https://github.com/IVRL/FG-NIC}).
\section{Computational complexity}
We show in Table~\ref{tab:mac_param} the number of Multiply–accumulate operation (MAC) and trainable parameters used in each network or module, highlighting the computational efficiency of our proposed approach.
\begin{table}[htbp]
  \centering
  \caption{
  \#MACs is the number of multiply–accumulate operations (in billion), and \#Params is the number of trainable parameters (in million).}
    \begin{tabular}{lcc}
    \toprule
    Network model & \#MACs (G) & \#Params (M) \\
    \hline
    \noalign{\smallskip}
    ResNet-50 (Classification) & 4.11       & 24.03 \\
    Ensemble module & 4.11       & 0.21 \\
    Our FG-NIC & 0.08       & 10.49 \\
    \bottomrule
    \end{tabular}
  \label{tab:mac_param}%
\end{table}%



\vspace{-10pt}
\section{Ablation study}
We run a series of experiments on Caltech-256 and conduct an in-depth analysis of the results to show the improvement of each module in our proposed method. {For the end-to-end setup, we train our FG-NIC and fidelity map estimator together. For the different fidelity map inputs and outputs, downsampling methods, and generally the ablation studies, we use the oracle setup with the single model.} The results are given in Table~\ref{tab:ablation} and support the statements and design decisions we make in our main manuscript. 
\begin{table}[htbp]
  \setlength\tabcolsep{2.5pt}
  \centering
  \caption{In-depth analysis and ablation study results (classification accuracy in \%). Bold numbers show the best accuracy and underlined numbers show the next best accuracy.}
    \begin{tabular}{ll ccccc}
    \toprule
    \multicolumn{2}{c}{\multirow{2}{*}{Methods}} & \multicolumn{5}{c}{Uniform degradation ($\sigma$)} \\
    & & 0.1 & 0.2 & 0.3 & 0.4 & 0.5 \\
    \hline
    \noalign{\smallskip}
    
    \multicolumn{2}{l}{\cellcolor{gray}Our FG-NIC: oracle + single} 
    & 80.45 & 78.14 & \bf{76.13} & \underline{74.06} & \underline{71.75} \\
    
    \multicolumn{2}{l}{\cellcolor{gray}Our FG-NIC: end-to-end + single} 
    & 80.57 & 75.21 & 68.22 & 60.01 & 51.03 \\
    \noalign{\smallskip}
    
    \multicolumn{2}{l}{\cellcolor{gray}Pretrained on clean test on restored} 
    & 77.99 & 67.06 & 53.00 & 38.08 & 25.53 \\
    \noalign{\smallskip}
    
    \multicolumn{2}{l}{\cellcolor{gray}Fidelity map input: restored}
    & 80.41 & \underline{78.15} & \bf{76.13} & \bf{74.12} & \bf{71.77} \\
    \noalign{\smallskip}
    
    \multicolumn{2}{l}{\cellcolor{gray}Fidelity map output: $\ell_2$ distance}
    & 80.46 & 78.13 & 76.08 & 73.90 & 71.55 \\            
    \multicolumn{2}{l}{\cellcolor{gray}Fidelity map output: cosine}
    & 79.45 & 75.07 & 69.27 & 62.56 & 55.18 \\
    \noalign{\smallskip}
    
    \multicolumn{2}{l}{\cellcolor{gray}Downsampling: bicubic}
    & \bf{80.69} & \bf{78.25} & \underline{76.09} & 73.81 & 71.43 \\
    
    \multicolumn{2}{l}{\cellcolor{gray}Downsampling: nearest neighbor}
    & 80.32 & 78.01 & 75.81 & 73.65 & 71.37 \\
    \noalign{\smallskip}
    
    \cellcolor{gray} & \cellcolor{gray}w/o spatial multiplication 
    & 80.17 & 77.40 & 74.89 & 72.34 & 69.49 \\

    \cellcolor{gray} & \cellcolor{gray}w/o spatial addition 
    & 80.09 & 76.93 & 73.82 & 70.47 & 67.08 \\
    
    \cellcolor{gray} & \cellcolor{gray}w/o channel multiplication 
    & 80.32 & 77.72 & 75.70 & 73.51 & 70.95 \\
    
    \multirow{-4}{*}{\cellcolor{gray}\makecell[l]{Ablation\\study}} & \cellcolor{gray}w/o channel concatenation 
    & \underline{80.55} & 77.55 & 74.68 & 71.77 & 68.63 \\
    \bottomrule
    \end{tabular}%
  \label{tab:ablation}%
\end{table}%

\bibliographystyle{IEEEtran}
\bibliography{references}